# DenseNet Models for Tiny ImageNet Classification

Zoheb Abai
IISc Bangalore
zohebabai@iisc.ac.in

Nishad Rajmalwar
UVCE Bangalore
nishad.rajmalwar@campusuvce.in

## Abstract

*In this paper, we present two image classification models on the Tiny ImageNet dataset. We built two very different networks from scratch based on the idea of Densely Connected Convolution Networks. The architecture of the networks is designed based on the image resolution of this specific dataset and by calculating the Receptive Field of the convolution layers. We also used some non-conventional techniques related to image augmentation and Cyclical Learning Rate to improve the accuracy of our models. The networks are trained under high constraints and low computation resources. We aimed to achieve top-1 validation accuracy of 60%; the results and error analysis are also presented.*

## 1. Introduction

There are several Image Classification competitions such as the ImageNet Large Scale Visual Recognition Challenge (ILSVRC), where several convolution neural network architectures and models are presented. In 2015, the ImageNet [1] challenge was won by the ResNet model with 152 layers which achieved a top-5 classification error of 3.57% [2]. In this paper, we present our approach for building classification models for a subset of ImageNet dataset called the Tiny ImageNet [4]. We did not use any pre-trained network available for the original ImageNet challenge. Instead, we built two different models from scratch taking inspiration from the DenseNet architecture [3]. We aimed to reach 60% validation accuracy with these custom models under several constraints and limited resources. Our biggest challenge was low computation power provided by Google Colab [5] free services. Google Colab provides only 12 hours of continuous computation time, after which the session needs to be reconnected. Also, we restricted our models from using any dense or fully connected layers, any dropout layers, 1x1 filters to increase the number of channels and more than 500 training epochs. We have used several innovative techniques tailored to this challenge in order to improve accuracy and reduce computation time.

## 2. Dataset

The Tiny ImageNet dataset [4] is a modified subset of the original ImageNet dataset [1]. Here, there are 200 different classes instead of 1000 classes of ImageNet dataset, with 100,000 training examples and 10,000 validation examples. The resolution of the images is just 64x64 pixels, which makes it more challenging to extract information from it. A glance at the images shows that it is difficult for the human eye to detect objects in some images.

## 3. DenseNet

We used the concept of DenseNet [3] to design our model architecture. The main challenge with very deep neural networks is the problem of vanishing gradients. This problem was first overcome by introducing residual networks which uses a shortcut connection to pass input from one block to another. In contrast to ResNet, DenseNet does not aggregate features through summation; instead, they are combined by concatenation. Thus the information is passed from one layer to all the subsequent layers ensuring better flow of information and gradients throughout the network.

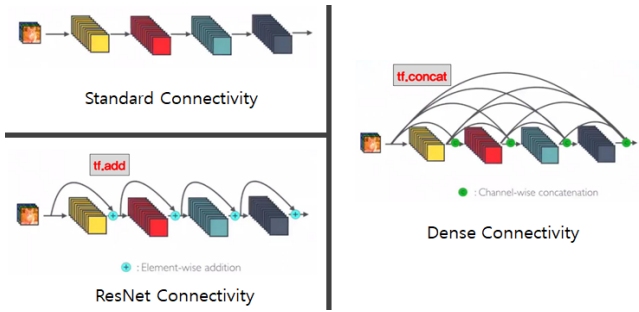

Figure 1. Standard vs ResNet vs DenseNet layers.
(Image Source: CVPR 2017 Presentation Slides)

## 4. Approach

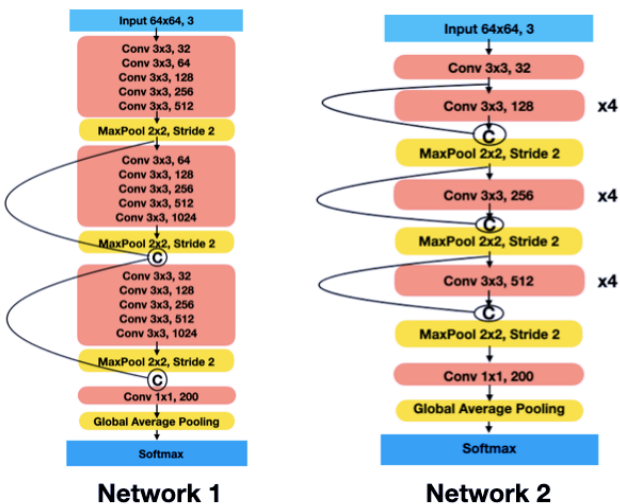

Figure 2. DenseNet Architectures of Networks

### 4.1 Receptive Field

For both of our networks, we implemented the number of layers in our model by first calculating the receptive field [6]. Throughout our model, we used (3x3) kernels with strides (1,1). So, for the first layer, the receptive field is (3x3) as each kernel convolutes over (3x3) pixels or 9 pixels at a time. Every such convolution operation decreases the spatial dimensions of the matrix by 2, thus increasing the receptive field of the network by 2. While at the MaxPooling layer, the spatial dimensions of the matrix are reduced by half, hence doubling the entire receptive field. The receptive field of the networks is calculated as shown in Figure 3:

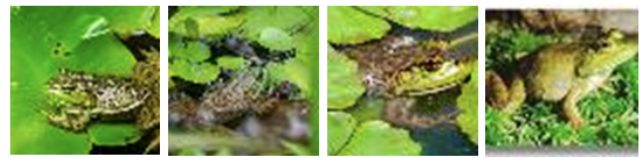

Figure 3. Receptive Field of Networks

As we reach the receptive field of original image size 64x64, we expect our model to detect the object in each image distinctly and be able to classify them. However, we have gone further, to a receptive field of more than double the original image size, so that network learns the background details too. Many images in our dataset are confusing due to background domination, so for those images, we want our models to understand the context in which our objects are found.

Figure 4. Images of Bullfrog

For example, in the class bullfrog as shown in Figure 4, we can observe that most of the images have a green background in addition to our object, i.e. bullfrog. We want our network to learn that in addition to our object.

### 4.2 Model Design

In the beginning, without much effort, we implemented the vanilla ResNet-34 and ResNet-50 models as our Network 1 and 2 respectively. For ResNet-34, we achieved an accuracy of 99.8% on the training set and 33.5% on the validation set with a batch size of 500 images running for 100 epochs. For ResNet-50 we achieved an accuracy of 49% on the training set and 26.2% on the validation set with a batch size of 512 images running for 50 epochs. Although this shows the learning power of ResNet models, we notice an early saturation in validation accuracy, due to the use of deep networks on shallow datasets [7].

Motivated by approaches explained in few Stanford 231n course project reports [8], we considered a modified approach for building a custom DenseNet model, having design elements of ResNet-18. The suitable reason behind the approach can be established from the fact that for deep networks on 64x64 images, the problem arises after first few layers where it starts working as an identity map of size 1x1 learning negligible with every additional layer. So we opt for a shallow network of around 15 convolution layers and wide enough to contain around 1000 channels before the output.

For Network 1 :
i. We built a custom architecture of 3 'bottleneck' blocks having 5 convolution layers of increasing channels and 1 MaxPooling layer at the end of each block.
ii. Just before applying concatenation, we feed the output of each block to space_to_depth function. This function ensures that the spatial dimensions of both the layers are equal before concatenation.
iii. We concatenate the skip connections from the output of each block with the output of the next block, preserving the information from both the blocks before being fed to the subsequent block.
iv. The final layers in the model have 1x1 convolution layer followed by a GlobalAveragePooling layer which averages the spatial dimensions of a matrix of any size. This layer gives us the ability to design a model which can take input image of any size. We shall use this fact to train our model better (to be explained further in Image Augmentation).

For Network 2 we modified the ResNet-18 architecture as follows:
i. We replaced the first convolution layer that initially consisted of 64 (7x7) filters with stride (2,2), by 32 (3x3) filters with stride (1,1) and removed the max pooling layer.
ii. We removed the 1st block consisting 4 convolution layers of 64 (3x3) filters.
iii. We removed the skip connections after every 2 convolution layers instead maintained it between every 4 convolution layers. We replaced the original add function in shortcuts with concatenation so that it preserves the channels from the previous block and not merge them. We added a Batch Normalization and ReLU activation layer after each shortcut.
iv. As per requirement of the project, we replaced final FC layers with 1x1 convolution layer for decreasing the number of channels to the required number of classes followed by a GlobalAveragePooling layer.

### 4.3 Image Augmentation

In order to artificially increase the amount of training data and avoid overfitting, we rely on image augmentation.

For Network 1, we used both direct and indirect approach towards implementing image augmentation. Firstly, as an indirect approach, we fed images with 32x32 resolutions for the beginning few epochs followed by 64x64 resolution images. Then we fed images with 16x16 resolutions for the next few numbers of epochs and finally fed the model with 64x64 resolutions images. The reason behind this was to feed low information scaled out data into the model till validation accuracy gets saturated so that it can learn a few features from these images. This technique gave us an increase of 3-4% in validation accuracy.

At 50% validation accuracy, we directly applied image augmentations from the imgaug library [11]. We manually provided the range of the parameters and randomly selected the number of augmentations (between 0 to all) from the list for each image. We used the following augmentations:
- Scale
- CoarseDropout
- Rotate
- Additive Gaussian Noise
- Crop and Pad

Applying this we got an increase of 9.5% on validation accuracy after an additional run for 150 epochs.

For Network 2, we kept default 64x64 image resolution as input size throughout our complete run. Instead of applying image augmentations after a certain number of iterations as in Network 1, we applied a random sequence of 11 transformations [8] to half of the dataset from the beginning of the model run using the imgaug library [11]. The intensity of each transformation was randomly determined within a specified range, but these range parameters were manually provided so that these transformed training images could closely represent the images in the validation dataset. The augmentations applied were :
- Horizontal Flip
- Vertical Flip
- Gaussian Blur
- Crop and Pad
- Scale
- Translate
- Rotate
- Shear
- Coarse Dropout
- Multiply
- Contrast Normalisation

## 4.4 Regularizers, Optimizers, Hyperparameters and Callbacks

Batch Normalization helps to normalise the inputs of the previous layer at each batch keeping the values in a comparable range with the mean equal to 0 and standard deviation equal to 1. This ensures the activations of our models do not get skewed at any one particular point and also increases the speed of computation. For both the networks we applied Batch Normalization after every convolution layer and then passed these values to the ReLU activation function. We used accuracy as our metric and categorical cross-entropy as our loss function.

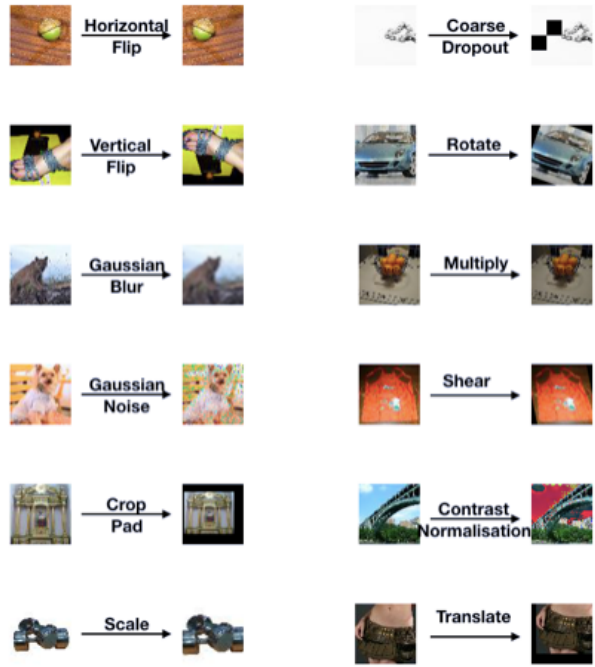

Figure 5. Examples of Image Augmentations

For Network 1, we chose the default Adam optimizer with a default learning rate of 1e-3 and it gave us good results but kept fluctuating with every epoch. We did not use any kernel regularizer nor did we change the default value of kernel initializer for this network. We used ReduceLRonPlateau as a callback function for reducing the learning rate if validation loss stagnates for 5 epochs.

For Network 2, we used L2 kernel regularizer with a lambda value of 2e-4 [8] with variance scaling kernel initializer [8] for every convolution layer. We used Adam optimizer with a learning rate of 1e-4 and epsilon value of 1e-8. Instead of decreasing the learning rate during plateaus as Network 1, we used the Cyclical Learning Rate method [12] where we vary the learning rate within a range of values for a specific number of iterations. This technique helps us in getting better validation accuracy in a fewer number of epochs. We used triangular2 rate policy, which halves the max learning rate with every cycle. As a modification to what is proposed in the original paper [12], we did not gradually reduce the range value of cycles and step sizes, after a certain number of cycles till the end. Instead, for the cycles where the validation accuracy does not show any increase than the previous range, we increase the learning rate range for those cycles and then lower the learning rate range for the subsequent cycle. Here, for the cycles of learning rate range 1e-7 to 6e-7 after the cycles of range 1e-6 to 6e-6, we did not see an improvement in validation accuracy, so we increased the learning rate range 1e-5 to 6e-5 for those cycles and then reduced the range 1e-7 to 6e-7 for the next cycle. This gave us an increase in the validation accuracy of 0.3%. This is summarised in Table 1 below.

Table 1

| Base_lr | Max_lr | Step-size in no. of epochs | Start Epoch | End Epoch | Max. Val. Accuracy |
|---|---|---|---|---|---|
| 1E-04 | 6E-04 | 6 | 0 | 24 | 53.26 |
| 1E-05 | 6E-05 | 6 | 24 | 48 | 60.86 |
| 1E-05 | 6E-05 | 4 | 48 | 72 | 61.74 |
| 1E-06 | 6E-06 | 2 | 72 | 84 | 62.43 |
| 1E-05 | 6E-05 | 2 | 84 | 96 | 61.83 |
| 1E-07 | 6E-07 | 2 | 96 | 108 | 62.73 |

As Google Colab [5] notebook is restricted to 12 hours of continuous use, for both the model runs we used a Model Checkpointer to save the model with best validation accuracy as we had to run over 100 epochs for each network.

## 5. Results

For Network 1 we trained our model with 17.9 Million parameters for 235 epochs with a batch size of 256 for 32x32 and 16x16 resolution images and a batch size of 64 for 64x64 resolution images. At first, we ran 32x32 resolution training images for 15 epochs and reached a validation accuracy of 25%. At this point, the model had already started overfitting, so we switched to 64x64 resolution images and ran it for 30 epochs which increased the validation accuracy to 48%. In order to train the model on low-resolution images and simulate scaled out augmentation, we trained it on 16x16 resolution images for just 10 epochs. We kept epochs low here because although we wanted the model to learn from low information images, we also wanted to avoid overfitting. We changed back to 64x64 resolution images and stuck to it for the rest of the training. After running it for 30 more epochs, our validation accuracy saturated around 48-49%, and now we implemented image augmentation. We performed some of the augmentations on every image and ran our model for another 150 epochs. Due to several augmentations, now the model trained much slower but had yet to begin overfitting. The gap between training and validation accuracy remained small, and we trained it until the validation accuracy saturated to around 59%. Here, the training accuracy reached 67%, so we did not train it further to avoid overfitting. We selected the best model from our training using model checkpoint which gave an accuracy of 59.5% on the validation dataset.

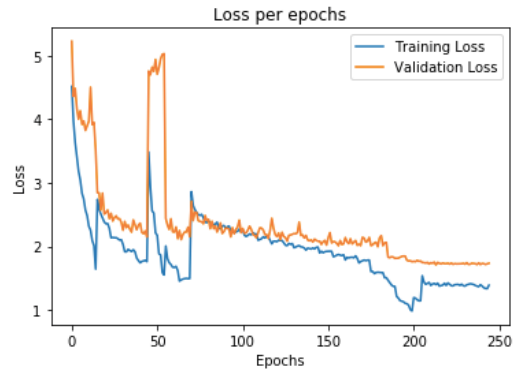

Figure 6. Loss curve of Network 1

For Network 2 we trained our model with 11.8 Million parameters for 108 epochs with a batch size of 128 images. As we can see from Figure 7, Cyclical Learning Rate helped us in early attainment of higher validation accuracy steeply bringing down the validation loss. It is also distinctly visible that after 70 epochs our validation loss reached a plateau for almost 10 epochs.That is when we applied higher Cyclical Learning Rate range due to which a sudden increase in loss is visible, after which the validation accuracy attains a new plateau although training accuracy continues decreasing. At the end of 108 epochs, we achieve a validation accuracy of 62.73% whereas our training accuracy remained at 68.11%, indicating a narrow gap again.

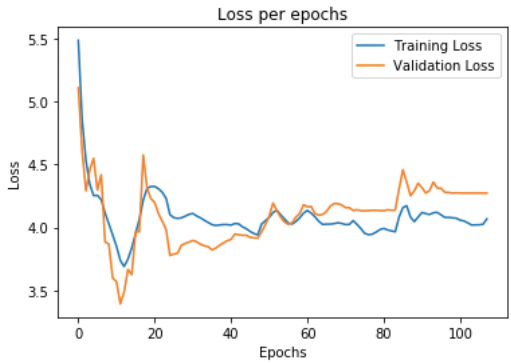

Figure 7. Loss curve of Network 2

Both the networks achieved validation accuracy higher than Shallow ConvNets [7] and ResNet-18 (with no Dropout) models [9] and are very close to state-of-the-art ResNet-18 (with various Dropout techniques) models [10].

## 6. Error Analysis

We visualise some interesting examples of top-1 errors made by our final model, as shown in the Figure 8.

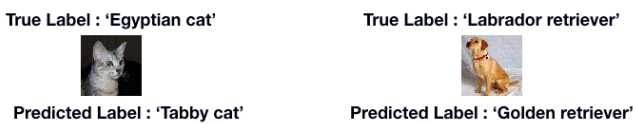

Figure 8. Examples of Mislabelled Classes

Overall, these errors are generally due to three main reasons [8]: the low resolution of the images, the misunderstanding of the primary entity in the image, and the confusion by similar items. Through manual inspection in both correctly and incorrectly classified images, we were able to draw the following detailed conclusions:

In the correctly classified images, we observed that there is a clear distinction between the object and the background. Images of every best classifying class tend to share same features within themselves such as colour, texture etc. Also, the model performs very well on detecting objects that cover the entire 64x64 resolution and has a minimal background. Though for some classes like 'flagpole', 'flagstaff', the model was able to detect zoomed in and zoomed out object. This could be because we fed low-resolution images and used the scaling augmentation.

In the incorrectly classified images, many images were misclassified due to the low resolution of the images. Some of them are even too hard for humans to differentiate. For example, an 'Egyptian cat' is misclassified as a 'Tabby cat' or a 'Labrador retriever' is misclassified as a 'Golden retriever'. Here, since there are not many details in the images, given the low resolution of 64x64, the model can detect the type of the object (such as a dog), but unable to further sub-classify it into the correct class (such as Labrador retriever and Golden retriever).

Another type of incorrectly classified images is due to the misunderstanding of the intent of the image [8]. The plate of fruits is classified as 'banana' simply because there is indeed a banana in it; however, the true label is 'orange'. A plate of food which is supposed to be categorised as 'meatloaf', is instead classified as a 'plate', which is perfectly reasonable. These kind of errors are impossible to fix. There is often more than one entity in a single image; our network interprets these images correctly but does not interpret what is indeed correct.

The last kind of misclassified images is due to the reason that the categories are too close to each other, sometimes there is even some overlap between each other. 'Convertible Cars' are misclassified as 'Sports Cars', but in reality, a large part of 'Convertible Cars' are also 'Sports Cars'.

In summary, although almost 30% of the images are not classified correctly, a large number of the errors are reasonable, some of them are even forgivable. Table 2 shows the top worst classified classes by our networks:

Table 2

| Class Name | Val. Accuracy |
|---|---|
| Plunger | 25% |
| Umbrella | 27.27% |
| Water Jug | 30% |
| Bucket | 30.43 |
| Wooden Spoon | 31.03% |
| Bannister | 31.71% |

After analysing our predictions, we separated correctly and incorrectly labelled images and oversampled the incorrectly labelled images three times the correctly labelled images. We also tried another approach by soft weighing the low precision classes. We expected our models to learn the incorrectly classified images better using these techniques; however, our model started overfitting, and we observed no improvement in validation accuracy.

## 7. Conclusions

We tried to overcome the challenge of Tiny ImageNet dataset under very high constraints and scarce computation resources. By designing two custom DenseNet models that are well suited for the challenge, we were able to achieve great results with Top-1 accuracy of 59.5% and 62.7% for our networks. We made our model as wide as possible and not much shallow to extract maximum features and prevent overfitting for our 64 resolution images. We used several data augmentation techniques for both models to increase our validation accuracy. To extract more information from the images, we used some promising techniques such as feeding varying resolutions of the input images and providing various ranges of Cyclic Learning Rate ranges. Implementing both of these techniques, we observed that for a few cycles it might have a short-term negative impact on the training but clearly shows long term beneficial effects evident from the jump of validation accuracy. We shall further endeavour on these techniques by applying them on other open datasets. At last, we manually inspected the misclassified images produced by our final model and gained valuable insights. For further studies, we shall fine-tune our model by custom tailoring the image augmentation parameters based on insights gained. We shall also focus on the architecture of the network and aim for achieving higher accuracy.

We want to acknowledge Rohan Shravan [13] for his constant support and guidance. Colab Notebook of the work is available at https://github.com/ZohebAbai/Tiny-ImageNet-Challenge

## References

[1] http://www.image-net.org/.
[2] K. He, X. Zhang, S. Ren, J. Sun, Deep Residual Learning for Image Recognition, arXiv:1512.03385v1, 2015
[3] G. Huang, Z. Liu, L. v. d. Maaten, K. Q. Weinberger, Densely Connected Convolutional Networks, arXiv:1608.06993v5, 2018
[4] https://tiny-imagenet.herokuapp.com/
[5] https://colab.research.google.com/
[6] H. Le, A. Borji, What are the Receptive, Effective Receptive, and Projective Fields of Neurons in Convolutional Neural Networks?, arXiv:1705.07049v2, 2018
[7] S. H. S. Basha, S R Dubey, V Pulabaigari, S Mukherjee, Impact of Fully Connected Layers on Performance of Convolutional Neural Networks for Image Classification, arXiv:1902.02771v2, 2019
[8] http://cs231n.stanford.edu/reports/2017/pdfs/ [pdfs: 12 & 930]
[9] M. S. Ebrahimi, H. K. Abadi, Study of Residual Networks for Image Recognition, arXiv:1805.00325v1, 2018
[10] R. Keshari, R. Singh, M. Vatsa, Guided Dropout, arXiv:1812.03965v1, 2018
[11] https://github.com/aleju/imgaug
[12] L. N. Smith, Cyclical Learning Rates for Training Neural Networks, arXiv:1506.01186v6, 2017
[13] https://in.linkedin.com/in/rohanshravan